\documentclass{article}

\usepackage{arxiv}

\usepackage{tabularx}
\usepackage{multirow}
\usepackage{array}
\usepackage{soul}
\usepackage{amsmath,amssymb,amsfonts}
\usepackage[ruled,vlined]{algorithm2e}
\usepackage{subcaption}

\usepackage[table,dvipsnames]{xcolor}

\usepackage[utf8]{inputenc} 
\usepackage[T1]{fontenc}    
\usepackage{hyperref} 
\usepackage{scalerel,stackengine}
\stackMath
\newcommand\reallywidehat[1]{%
\savestack{\tmpbox}{\stretchto{%
  \scaleto{%
    \scalerel*[\widthof{\ensuremath{#1}}]{\kern-.6pt\bigwedge\kern-.6pt}%
    {\rule[-\textheight/2]{1ex}{\textheight}}
  }{\textheight}%
}{0.5ex}}%
\stackon[1pt]{#1}{\tmpbox}%
}

\usepackage{graphicx}
\usepackage[export]{adjustbox}
\newcommand{\foo}{\hspace{-2.3pt}\textcolor{blue}{$\bullet$} \hspace{5pt}}
\usepackage{colortbl}
\usepackage{paralist}
\usepackage{bbm}
\usepackage{url}            
\usepackage{booktabs}       
\usepackage{amsfonts}       
\usepackage{nicefrac}       
\usepackage{microtype}      
\usepackage{lipsum}	
\usepackage{pifont}

\usepackage{tikz}
\usetikzlibrary{shapes.geometric}

\usepackage{relsize}
\usepackage{mathtools}

\usepackage{multirow}
\usepackage{multicol}
\usepackage{paralist}
\usepackage{float}
\usepackage[numbers]{natbib}

\title{When \textit{Neutral} Summaries are not that \textit{Neutral}: Quantifying Political Neutrality in LLM-Generated News Summaries}

\author{
Supriti Vijay\thanks{Supriti Vijay and Aman Priyanshu are equal contribution first authors.} \\
  \small{Carnegie Mellon University}\\
  \texttt{supritiv@andrew.cmu.edu} \\
  \And
  Aman Priyanshu$^*$\\
  \small{Carnegie Mellon University}\\
  \texttt{apriyans@andrew.cmu.edu} \\
\And
Ashique R. KhudaBukhsh \\
  \small{Rochester Institute of Technology}\\
  \texttt{axkvse@rit.edu} \\
}

\begin{document}
\maketitle
\begin{abstract} 
In an era where societal narratives are increasingly shaped by algorithmic curation, investigating the political neutrality of LLMs is an important research question. This study presents a fresh perspective on quantifying the political neutrality of LLMs through the lens of abstractive text summarization of polarizing news articles. We consider five pressing issues in current US politics: \textit{abortion}, \textit{gun control/rights}, \textit{healthcare}, \textit{immigration}, and \textit{LGBTQ+ rights}. Via a substantial corpus of 20,344 news articles, our study reveals a consistent trend towards pro-Democratic biases in several well-known LLMs, with gun control and healthcare exhibiting the most pronounced biases (max polarization differences of -9.49\% and -6.14\%, respectively). Further analysis uncovers a strong convergence in the vocabulary of the LLM outputs for these divisive topics (55\% overlap for Democrat-leaning representations, 52\% for Republican). Being months away from a US election of consequence, we consider our findings important. 
\end{abstract}


\section{Introduction}

Political polarization in the US is a widely studied problem across several disciplines~\citep{poole1984polarization, layman2010party,mccarty2016polarized, baldwin2016past,mcconnell2017research,khudabukhsh2021we,simchon2022troll}. Prior behavioral studies indicate that negative views towards the political \textit{other} have influenced outcomes in as diverse settings as allocating scholarship funds~\citep{iyengar2015fear}, mate selection~\citep{huber2017political}, employment decisions~\citep{gift2015does}, and even who finds what offensive~\citep{weerasooriya-etal-2023-vicarious}! In an era where societal narratives are increasingly shaped by algorithmic curation~\citep{stephens2017everybody,hofmann2024dialect}, and as large language models (LLMs) are increasingly being leveraged for a wide range of tasks~\citep{castelvecchi2024deepmind,fang2023recruitpro,ziems2023large,thirunavukarasu2023large} and integrated into technological applications~\citep{mobilePhoneAI}, investigating the political neutrality of large language models is an important research question.

\textbf{\textit{How do we quantify the political neutrality of large language models?}} This paper presents a fresh perspective on quantifying political neutrality through the lens of abstractive summarization of news articles on hot-button issues. Text summarization, a pivotal natural language processing task researched for decades~\citep{dorr2003hedge,DBLP:conf/naacl/ChopraAR16,DBLP:conf/emnlp/RushCW15, DBLP:journals/corr/RanzatoCAZ15,DBLP:conf/acl/SeeLM17,stiennon2020learning}, inherently demands fidelity to the source material. That said, our study reveals that it is fairly straightforward to construct politically colored news summaries catering to a specific audience leveraging cutting-edge LLMs. Unsurprisingly, such summaries highlight the unmistakable partisan liberal and conservative talking points. What is particularly intriguing is the (lack of) neutrality of the default news summaries -- the ones generated without any specified political orientation. Via a substantial corpus of 20,344 news summaries on five hot-button issues (\textit{abortion}, \textit{gun control/rights}, \textit{immigration}, \textit{healthcare}, and \textit{LTBTQ+ rights}) over which multiple US elections have been contested, our study reveals that default news summaries generated by several cutting-edge LLMs show liberal bias -- in short, the default is not (politically) neutral.

Our contributions are the following.\\
\foo\textit{\textbf{Framework:}} We present a novel framework to quantify political neutrality in leading open-source or freely available large language models. We prompt each model to produce summaries of polarizing news articles from three distinct perspectives -- neutral (default); Democrat-aligned; and Republican-aligned. Extending prior literature in quantifying cross-corpora differences~\citep{DuttaPoliceIJCAI2022}, we propose a metric to quantify political neutrality (described in detail in Section~\ref{Sec:QuantifyingNeutrality}) of large language models.\\
\foo\textit{\textbf{Social:}} Our study reveals that existing high-performance LLMs exhibit liberal bias while processing political content. The extent of bias varies across LLMs and issues. We analyze divergence in vocabulary and thematic representation between the summaries to understand how LLMs frame topics for varied perceived audience ideologies. We investigate algorithmic monoculture~\citep{bommasani2022picking} and evaluate whether biases and thematic choices remain consistent across different LLMs.\\
\foo\textit{\textbf{Resource:}} We release a dataset of 20,344 news articles and their corresponding left-leaning, right-leaning, and neutral summaries obtained from four large language models\footnote{Publicly available upon acceptance}. This dataset can be valuable for political alignment research.\\ 

\section{Related Work}

\subsection{Bias in LLMs}

The investigation into biases within large language models (LLMs) has garnered increasing attention due to their significant societal impact. Studies have delved into measuring political biases and predicting individual users' ideologies (\cite{colleoni2014echo, makazhanov2013predicting, preotiuc-pietro-etal-2017-beyond}) as well as analyzing news articles (\cite{li-goldwasser-2019-encoding, feng2022kgap, liu-etal-2022-politics,zhang-etal-2022-kcd}).

Notably, limited research has explored political biases within generative LLMs (\cite{liu2021mitigating,feng-etal-2023-pretraining,khorramrouz2023toxicity,motoki2023more}). \cite{liu2021mitigating} introduced metrics to gauge political bias in GPT-2 generation, proposing a reinforcement learning framework to mitigate such biases. \cite{feng-etal-2023-pretraining} examined LLMs using political science-grounded prompts to measure their ideological positions on social and economic values, alongside investigating the influence of political biases in pretraining data on the models' political leanings and their performance on downstream tasks. \cite{khorramrouz2023toxicity} extensively documented biases in models like PaLM-2, elucidating the challenges these biases pose to ethical AI development. \citet{motoki2023more} uncovered ideological slants within LLM-generated content, highlighting an intrinsic polarization mirroring societal divisions. However, these studies primarily focus on binary classifications of bias or qualitative assessments, leaving a gap in quantitative, comparative analysis across multiple LLMs and ideological spectrums. Our work aims to provide a comprehensive understanding of biases in LLMs, bridging the gap between qualitative assessments and quantitative analyses across various ideological spectrums and model architectures. Additionally, we explore political polarity disambiguation through the lens of classification tasks, inspired by \cite{DuttaPoliceIJCAI2022}, as a means of quantifying bias in LLMs. This allows us to compare the default ideological leanings across various LLMs in a systematic manner, further providing insights into the nuanced nature of political biases within these models and their implications.

\subsection{Algorithmic Monoculture}

The concept of algorithmic monoculture, while discussed in the context of software engineering and digital platforms \cite{peng2023monoculture, bommasani2022picking}, has only recently been applied to the study of LLMs \cite{levent2023model,bommasani2022picking,khorramrouz2023toxicity}. This concept posits that homogeneity in algorithmic decision-making processes, especially in content generation and moderation, can lead to a narrowing of perspectives and reinforcement of biases. While \cite{levent2023model, hofmann2024dialect, priyanshu2024silentcurriculumdoesllm} have touched upon the risks of monocultural tendencies in AI, there is a scarcity of empirical research investigating these risks across the spectrum of available LLMs, especially concerning their handling of politically polarized content. In this paper, we explore the presence of algorithmic monoculture across various LLMs and highlight its potential effects on exacerbating the spread of biased viewpoints.

\section{Experimental Setting}

This section describes the approach we took to examine how political biases are embedded and propagated in large language models (LLMs), as well as the potential for a uniformity of thought---what is termed algorithmic monoculture---among different LLM architectures. Our research evaluates how LLMs generate summaries of politically contentious news articles. Through these experiments, the goal is to measure the political leanings embedded within these models, pinpointing any biases and quantifying their relative impacts with respect to the two largest political parties in the US---Democrats and Republicans. This assessment is essential for determining the appropriateness of using these models to develop political narratives, perspectives, and summaries for news articles. 

\begin{figure}[h!]
    \centering
    \includegraphics[width=0.455\textwidth]{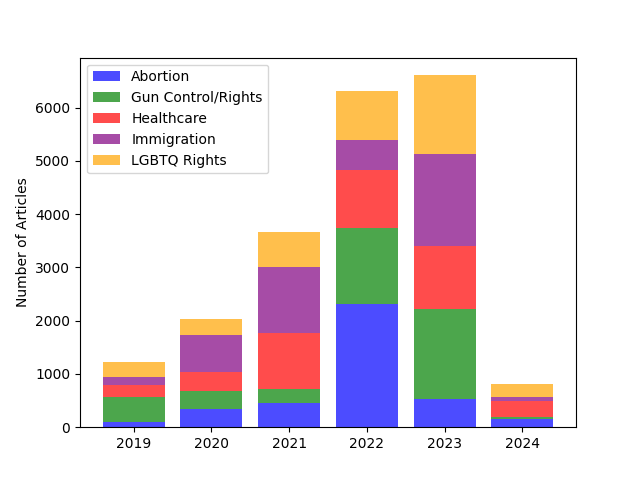}
    \caption{This image depicts the publication years of articles scraped for our study, spanning from 2019 to 2024. Each bar represents the quantity of articles from a specific year included in the analysis.}
    \label{fig:yeardistribution}
\end{figure}

\subsection{Dataset Collection}

\begin{table}[h!]
\centering
\scriptsize
\begin{tabular}{|l|l|l|l|}
\hline
\multicolumn{1}{|c|}{\textbf{Topic}} & \textbf{\begin{tabular}[c]{@{}l@{}}Number of\\ Articles\end{tabular}} & \textbf{\begin{tabular}[c]{@{}l@{}}Average Number of\\ Words / Article\end{tabular}} & \textbf{\begin{tabular}[c]{@{}l@{}}Average Number of\\ Sentences / Article\end{tabular}} \\ \hline
Abortion & 3,896 & 1202.36 & 64.75 \\ \hline
Gun control/rights & 4,214 & 1134.64 & 54.77 \\ \hline
Healthcare & 4,238 & 1397.88 & 68.15 \\ \hline
Immigration & 4,115 & 1082.53 & 56.74 \\ \hline
LGBTQ+ rights & 3,881 & 1280.79 & 56.41 \\ \hline
\end{tabular}
\caption{Overview of dataset composition by topic, including the total number of news articles scraped, average words, and sentences per article for key hot-button issues/polarizing topics (abortion, gun control/rights, healthcare, immigration, LGBTQ+ rights) from 2019 to present (2024).}
\label{tab:dataset}
\end{table}

We collect a diverse dataset of news articles from 2019 to the present year (2024), on five hot-button issues in the US: \textit{abortion}, \textit{gun control/rights}, \textit{healthcare}, \textit{immigration}, and \textit{LGBTQ+ rights}. These issues have broad overlap with prior literature on the US political divide~\citep{mouw2001culture,tesler2012spillover,miller2019americans,hout2021immigration,edenborg2023traditional} with many receiving individual focus within the NLP community~\citep{demszky-etal-2019-analyzing,mendelsohn2020framework,ramesh-etal-2022-revisiting}. Figure~\ref{fig:yeardistribution} illustrates the distribution of articles by their publication years. We prioritize including the most recent articles. Due to this design consideration, the majority of the articles scraped are from 2022 and 2023. We use GNews (Google-News) and scrape relevant articles on the aforementioned issues. Table~\ref{tab:dataset} lists summary statistics of this dataset. 

\begin{table}[h!]
\scriptsize
\centering
\begin{tabular}{|l|llll|}
\hline
\multicolumn{1}{|c|}{\multirow{2}{*}{\textbf{Models}}} & \multicolumn{4}{c|}{\textbf{Average number of words per summary}} \\ \cline{2-5} 
\multicolumn{1}{|c|}{} & \multicolumn{1}{l|}{\textbf{Pro-Democratic}} & \multicolumn{1}{l|}{\textbf{Pro-Republican}} & \multicolumn{1}{l|}{\textbf{Default/}} & \textbf{Aggregate} \\
\multicolumn{1}{|c|}{} & \multicolumn{1}{l|}{} & \multicolumn{1}{l|}{} & \multicolumn{1}{l|}{\textbf{Neutral}} &  \\ \hline
LLaMA-7B & \multicolumn{1}{l|}{164.19} & \multicolumn{1}{l|}{169.23} & \multicolumn{1}{l|}{154.56} & 162.66 \\ \hline
Vicuna-7B & \multicolumn{1}{l|}{139.12} & \multicolumn{1}{l|}{138.27} & \multicolumn{1}{l|}{138.61} & 138.67 \\ \hline
Mistral-7B & \multicolumn{1}{l|}{155.55} & \multicolumn{1}{l|}{155.58} & \multicolumn{1}{l|}{150.81} & 153.98 \\ \hline
PaLM-2 & \multicolumn{1}{l|}{101.42} & \multicolumn{1}{l|}{97.52} & \multicolumn{1}{l|}{90.48} & 96.47 \\ \hline
\end{tabular}
\caption{Presents a comparative analysis of the summarization outcomes across five large language models (LLaMA-7b, Mistral-7B, Vicuna-7B, and PaLM-2), detailing the average word count for summaries aligned with Pro-Democratic \& Pro-Republican perspectives, and control/neutral summaries. This table encapsulates the core quantitative analysis of summaries examined for ideological bias across politically polarizing topics.}
\label{tab:summary_description}
\end{table}

\subsection{Partisan Summary Extraction}
We consider four well-known large language models: \texttt{LLaMA-7B}~\citep{touvron2023llama}, \texttt{Mistral-7B}~\citep{jiang2023mistral}, \texttt{Vicuna-7B}~\citep{chiang2023vicuna}, and \texttt{PaLM 2}~\citep{anil2023palm}
. Each model is tasked with producing summaries for the aggregated content. For each article, we prompt the models to create three distinct summaries: one with a specified alignment towards Pro-Democratic viewpoints, another with a Pro-Republican alignment (the specific prompts for these alignments are detailed in the Appendix to be released in Supplementary Materials), and a third, which serves as a control or default summary. This methodology aims to assess the inherent biases in the models' summarization processes without explicit directional cues for two of the summaries and to compare these against the control. Overall, this process yields 250k+ summaries, providing a comprehensive basis for analysis. Table~\ref{tab:summary_description} presents the statistics of the produced summaries. 

\subsection{Diverging Vocabulary Analysis}
\label{sec:divvocab}

To further delve into the ideological biases inherent in the generated summaries, our approach included a comprehensive diverging vocabulary analysis. This analysis was designed to quantitatively identify and compare the frequency of specific lexical choices---words/phrases (tokens)---across two distinct sets of summaries, characterized as Democrat-leaning ($\mathcal{D}$) and Republican-leaning ($\mathcal{R}$). For each $\mathcal{D}$ and $\mathcal{R}$, we compute the respective unigram distributions $\mathcal{N}_p^D$ and $\mathcal{N}_p^R$. Next, for each token $t$, we compute the token bias scores $\mathcal{B}(t) = \mathcal{N}_p^R(t) - \mathcal{N}_p^D(t)$. Tokens with a positive $\mathcal{B}(t)$ are indicative of a Republican bias, whereas a negative $\mathcal{B}(t)$ suggests a Democrat bias.

\subsection{Political Ideology Polarization Quantification}\label{Sec:QuantifyingNeutrality}

In order to quantify political neutrality of the summaries, we need to identify if the pro-Republican or the pro-Democratic summaries are \textit{closer} to the neutral summaries. 
Quantifying differences between text corpora is a well-studied problem~\citep{khudabukhsh2021we,pillutla2021mauve,DuttaPoliceIJCAI2022}. We extend a text classification-based approach that uses classification accuracy as a proxy for text separability outlined in~\cite{DuttaPoliceIJCAI2022}. For two text corpora, $\mathcal{D}_1$ and $\mathcal{D}_2$, a classsifer is trained to predict the source dataset. Intuititevely, if the datasets are easily distinguishable, the classification accuracy will be high on a balanced test set. If not, the accuracy will be close to chance.   

In our approach, we construct two distinct SentenceTransformer-based \citep{wang2020minilm} classifiers for each political party: (1) one tailored to identify disparities between default (neutral) summaries and summaries with a Democratic inclination, and the other (2) to identify disparities between default (neutral) summaries and summaries with a Republican inclination. We adopt a 5-fold cross-validation method, reinforcing the reliability and reproducibility of our findings, details of which can be found in Appendix to be released in Supplementary Materials.

This analytical approach assesses the ideological divergence from a purportedly neutral standpoint, as represented by the default summaries. Let us define the effectiveness of a classifier in distinguishing between neutral and politically-aligned summaries as \(\textit{diff}(\mathcal{D}_{\textit{Democrat}}, \mathcal{D}_{\textit{Neutral}})\) for Democrat and \(\textit{diff}(\mathcal{D}_{\textit{Republican}}, \mathcal{D}_{\textit{Neutral}})\) for Republican. Drawing on previous studies that use classification accuracy as a proxy for polarization, we adopt accuracy as our metric for assessing ideological bias \citep{DuttaPoliceIJCAI2022}. The classifier's accuracy, \(\textit{diff}(\mathcal{D}_{\textit{Political Ideology}}, \mathcal{D}_{\textit{Neutral}})\), reflects its capacity to differentiate between these groups, with values closer to 1 indicating clear distinction and values near 0.5 highlighting an inability to distinguish effectively. Therefore, values of \(\textit{diff(.)}\) close to 0.5 would imply a significant overlap between neutral and politically aligned summaries, suggesting minimal ideological bias from the neutral standpoint. To encapsulate the degree of polarization, we introduce a quantification measure, the Polarization Index ($\mathcal{P}$ ), calculated as:
\[
\mathcal{P} = \textit{diff}(\mathcal{D}_{\textit{Democrat}}, \mathcal{D}_{\textit{Neutral}}) - \textit{diff}(\mathcal{D}_{\textit{Republican}}, \mathcal{D}_{\textit{Neutral}})
\]
This index measures the degree to which the default summaries diverge towards a particular ideological bias. A positive $\mathcal{P}$ indicates a bias towards Republican views, whereas a negative $\mathcal{P}$ suggests a bias towards Democratic views. This measure allows us to capture the nuanced ideological tilts and biases inherent in the LLM-generated summaries, offering a refined lens through which we assess the degree of alignment or deviation from a truly neutral or balanced depiction of political discourse.

\subsection{Investigating Algorithmic Monoculture}

\cite{bommasani2022picking} posit that given that high-performance LLMs are trained on comparable datasets and architecture, for the same entity group, the content outcome across LLMs may bear a strong resemblance. Employing this as a baseline, this work inspects algorithmic monoculture in two key ways. 

\textbf{Vocabulary-based monoculture analysis:} Our analysis extends to examining whether the top \(N = 20\) tokens with the greatest divergence for either ideology, as identified by the token bias metric $\mathcal{B}(t)$ for each model, demonstrate recurring patterns across them. Let \(T_{M}\) denote the set of top \(N\) tokens for each LLM \(M\), where $\mathcal{M} \in \{\texttt{LLaMA-7b}, \texttt{Mistral-7B}, \texttt{Vicuna-7B}, \texttt{PaLM 2} \}$. This comparative examination across models/LLMs, particularly concerning their handling of sensitive topics, enables us to assess whether the ideological framing exhibits consistency.

We define a consistency index $\mathcal{CI}$ to quantify the extent of algorithmic monoculture across models in percentage terms: $\mathcal{CI} = \frac{1}{|\mathcal{M}|(|\mathcal{M}|-1)} \sum_{{\mathcal{M}_i, \mathcal{M}_j \in \mathcal{M}}} |\mathcal{T}_{\mathcal{M}_i} \cap \mathcal{T}_{\mathcal{M}_j}| \times 100$, where $|\mathcal{T}_{\mathcal{M}_i} \cap \mathcal{T}_{\mathcal{M}_j}|$ is the number of tokens that appear in the top \(N\) lists of both models $\mathcal{M}_i$  and $\mathcal{M}_j$. A higher $\mathcal{CI}$ would suggest a greater degree of consensus in the ideological biases across different LLMs, indicating a potential algorithmic monoculture. Conversely, a lower $\mathcal{CI}$ would point to a more diverse representation of ideological stances.

\textbf{Classifier-based monoculture analysis:} To delve into the phenomenon of algorithmic monoculture, we employed SentenceTransformer-based classifiers, initially trained on summaries from a specific LLM (denoted as \(\mathcal{M}_{\textit{source}}\), e.g., \texttt{LLaMA-7B}), and applied them to the summaries generated by the other LLMs (denoted as \(\mathcal{M}_{\textit{target}}\), e.g., \texttt{PaLM-2}). This cross-LLM analysis, symbolized as \(\mathcal{A}(\mathcal{M}_{\textit{source}}, \mathcal{M}_{\textit{target}})\), facilitated an examination of the extent to which different models exhibit shared biases or employ analogous linguistic strategies in their production of politically inclined summaries.

\section{Result}

\subsection{Vocabulary Divergence}
\renewcommand{\tabcolsep}{1mm}
\begin{table*}[h!]
\centering
\small
\begin{tabular}{|cc|cc|cc|cc|cc|}
\hline
\multicolumn{2}{|c|}{\textbf{Abortion}} & \multicolumn{2}{c|}{\textbf{Gun Control/Rights}} & \multicolumn{2}{c|}{\textbf{Healthcare}} & \multicolumn{2}{c|}{\textbf{Immigration}} & \multicolumn{2}{c|}{\textbf{LGBTQ+}} \\ \hline
\multicolumn{1}{|c|}{\textbf{Democrat}} & \textbf{Republican} & \multicolumn{1}{c|}{\textbf{Democrat}} & \textbf{Republican} & \multicolumn{1}{c|}{\textbf{Democrat}} & \textbf{Republican} & \multicolumn{1}{c|}{\textbf{Democrat}} & \textbf{Republican} & \multicolumn{1}{c|}{\textbf{Democrat}} & \textbf{Republican} \\ \hline
\multicolumn{1}{|c|}{\cellcolor{blue!15}\textit{right}} & \cellcolor{red!15}\textit{mother} & \multicolumn{1}{c|}{\cellcolor{blue!15}\textit{gun}} & \cellcolor{red!15}\textit{find} & \multicolumn{1}{c|}{\cellcolor{blue!15}\textit{health}} & \cellcolor{red!15}\textit{limited} & \multicolumn{1}{c|}{\cellcolor{blue!15}\textit{advocate}} & \cellcolor{red!15}\textit{security} & \multicolumn{1}{c|}{\cellcolor{blue!15}\textit{lgbtq+}} & \cellcolor{red!15}\textit{allow} \\ \hline
\multicolumn{1}{|c|}{\cellcolor{blue!15}\textit{reproductive}} & \cellcolor{red!15}\textit{provider} & \multicolumn{1}{c|}{\cellcolor{blue!15}\textit{strict}} & \cellcolor{red!15}\textit{abide} & \multicolumn{1}{c|}{\cellcolor{blue!15}\textit{access}} & \cellcolor{red!15}\textit{bear} & \multicolumn{1}{c|}{\cellcolor{blue!15}\textit{comprehensive}} & \cellcolor{red!15}\textit{fence} & \multicolumn{1}{c|}{\cellcolor{blue!15}\textit{right}} & \cellcolor{red!15}\textit{believe} \\ \hline
\multicolumn{1}{|c|}{\cellcolor{blue!15}\textit{access}} & \cellcolor{red!15}\textit{advocate} & \multicolumn{1}{c|}{\cellcolor{blue!15}\textit{violence}} & \cellcolor{red!15}\textit{news} & \multicolumn{1}{c|}{\cellcolor{blue!15}\textit{care}} & \cellcolor{red!15}\textit{high} & \multicolumn{1}{c|}{\cellcolor{blue!15}\textit{reform}} & \cellcolor{red!15}\textit{american} & \multicolumn{1}{c|}{\cellcolor{blue!15}\textit{community}} & \cellcolor{red!15}\textit{belief} \\ \hline
\multicolumn{1}{|c|}{\cellcolor{blue!15}\textit{women}} & \cellcolor{red!15}\textit{stance} & \multicolumn{1}{c|}{\cellcolor{blue!15}\textit{control}} & \cellcolor{red!15}\textit{issue} & \multicolumn{1}{c|}{\cellcolor{blue!15}\textit{bill}} & \cellcolor{red!15}\textit{politic} & \multicolumn{1}{c|}{\cellcolor{blue!15}\textit{measure}} & \cellcolor{red!15}\textit{important} & \multicolumn{1}{c|}{\cellcolor{blue!15}\textit{advocate}} & \cellcolor{red!15}\textit{religious} \\ \hline
\multicolumn{1}{|c|}{\cellcolor{blue!15}\textit{health}} & \cellcolor{red!15}\textit{conservative} & \multicolumn{1}{c|}{\cellcolor{blue!15}\textit{shooting}} & \cellcolor{red!15}\textit{rare} & \multicolumn{1}{c|}{\cellcolor{blue!15}\textit{universal}} & \cellcolor{red!15}\textit{cost} & \multicolumn{1}{c|}{\cellcolor{blue!15}\textit{call}} & \cellcolor{red!15}\textit{secure} & \multicolumn{1}{c|}{\cellcolor{blue!15}\textit{continue}} & \cellcolor{red!15}\textit{traditional} \\ \hline
\end{tabular}
\caption{Comparative analysis of the top-5 tokens as per the $\mathcal{B}(t)$ score, which represent the increased usage in the respective respective subgroups of $\mathcal{R}$ and $\mathcal{D}$, for the topics of Abortion, Gun Control/Rights, Healthcare, Immigration, and LGBTQ+, reflecting Democrat and Republican ideological biases in language model summaries. The table illustrates words with the most negative $\mathcal{B}(t)$ for democrats and most positive $\mathcal{B}(t)$ for republicans, highlighting the nuanced ideological biases within these sensitive areas.}
\label{tab:divergingvocab}
\end{table*}

Our analysis revealed distinct vocabulary biases within summaries aligned to Democrat and Republican viewpoints across the five key topics: \textit{abortion}, \textit{gun control/rights}, \textit{healthcare}, \textit{immigration}, and \textit{LGBTQ+ rights}. Employing the token bias score discussed in Section~\ref{sec:divvocab} $\mathcal{B}(t) = \mathcal{N}_p^R(t) - \mathcal{N}_p^D(t)$, we identified the top five divergent terms for each topic, as shown in Table~\ref{tab:divergingvocab}. The table details the top-5 tokens at either extremity of this score (indicating increased usage in the respective sub-corpus), offering a snapshot of the lexical choices that differentiate Democrat from Republican narratives as per these models. These findings serve to quantitatively illustrate the lexical patterns and ideological biases inherent in the generated summaries. For a more detailed analysis and to gain a better understanding, we have listed the top-5 tokens for each topic across every LLM in Appendix(to be released in Supplementary Materials).

Table~\ref{tab:divergingvocab} indicates that when LLMs are tasked with generating pro-Democratic news summaries, key liberal talking points, such as reproductive rights~\citep{mouw2001culture}, stricter gun controls~\citep{miller2019americans}, or universal healthcare~\citep{tesler2012spillover} get considerably emphasized. In contrast, when these very same LLMs are instructed to generate pro-Republican news summaries, key conservative talking points such as high health care cost~\citep{tesler2012spillover}, border security~\citep{chacon2006unsecured}, and traditionalist values towards sexual orientation~\citep{edenborg2023traditional} emerge.

\subsubsection{Monoculture Analysis: Do LLMs use the same vocabulary?}

\begin{figure}[h!]
    \centering
    \begin{subfigure}[b]{0.49\textwidth}
        \centering
        \includegraphics[width=\textwidth]{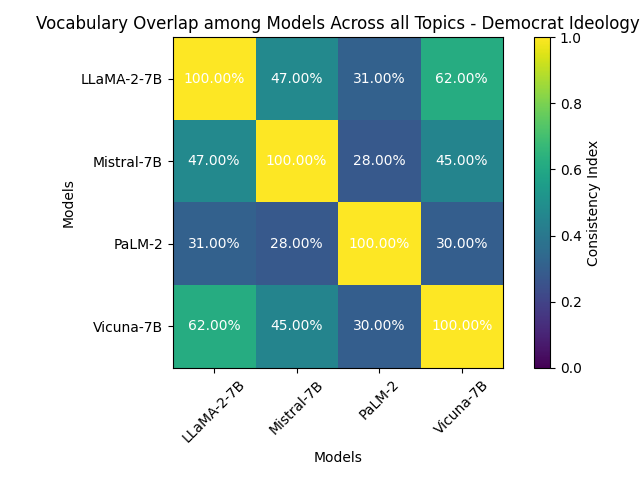}
        \caption{\scriptsize Democrat-leaning vocabulary overlap}
        \label{fig:democratvocabmonoculture}
    \end{subfigure}
    \hfill
    \begin{subfigure}[b]{0.49\textwidth}
        \centering
        \includegraphics[width=\textwidth]{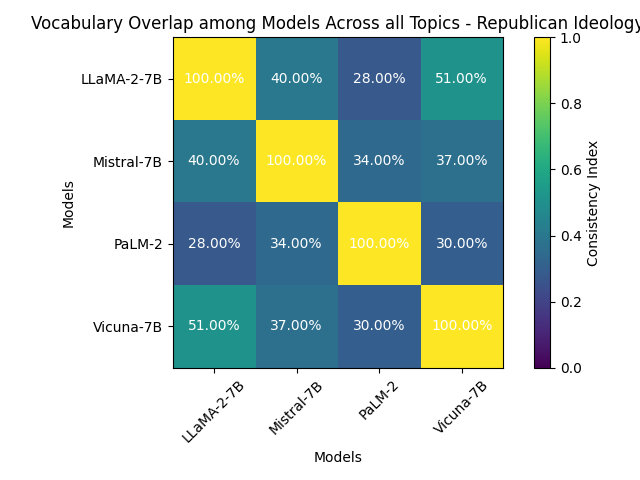}
        \caption{\scriptsize Republican-leaning vocabulary overlap}
        \label{fig:republicanvocabmonoculture}
    \end{subfigure}
    \caption{Heatmaps illustrating the consistency index $\mathcal{CI}$ across various large language models (LLMs) for both Democrat and Republican-leaning vocabularies. These visualizations serve as a measure of ideological congruence, revealing patterns of uniformity, if any, in how different LLMs linguistically frame their opinions.}
    \label{fig:vocabmonoculture}
\end{figure}

Our investigation into the concept of algorithmic monoculture, through the lens of ideological bias in language model outputs, reveals intriguing patterns of diverging vocabulary overlap across different models. Utilizing the consistency index $\mathcal{CI}$ to quantify the extent of shared ideological vocabulary, we found that the overall average $\mathcal{CI}$ for Democrat-leaning vocabulary stands at 55.37\% (40.5\% for non-diagonal mean---here we consider diagonal as top-left to bottom-right, which represents overlap over the same LLM outputs), slightly higher than the Republican-leaning vocabulary $\mathcal{CI}$ of 52.49\% (36.65\% for non-diagonal mean). This suggests a slightly more pronounced consensus among models on Democrat-biased terms. The maximum non-diagonal vocabulary overlap for Democrat-leaning terms was observed between \texttt{LLaMA-2} and \texttt{Vicuna} at 62\%, with the minimum at 28\% between \texttt{Mistral} and \texttt{PaLM-2}, indicating a variability in ideological representation with a standard deviation of \(\pm\)27.82\%. Similarly, for Republican-biased terms, the greatest overlap was between \texttt{LLaMA-2} and \texttt{Vicuna} at 51\%, and the least was between \texttt{LLaMA} and \texttt{PaLM-2} at 28\%, with a standard deviation of \(\pm\)28.19\%. These statistics, illustrated in our heatmap visualizations in Figure~\ref{fig:vocabmonoculture}, underscore the varying degrees of ideological bias present across different large language models (LLMs) and point towards a certain level of algorithmic monoculture, as evidenced by an average overlap of more than 50\% (or more than 35\% for non-diagonal) of the tokens for either ideology. 

\subsection{Polarization Quantification}

In exploring the polarization of political ideologies as represented by LLMs, our analysis leveraged the differential effectiveness of SentenceTransformer-based classifiers, denoted as \(\textit{diff}(\mathcal{D}_{\textit{Democrat}}, \mathcal{D}_{\textit{Neutral}})\) and \(\textit{diff}(\mathcal{D}_{\textit{Republican}}, \mathcal{D}_{\textit{Neutral}})\), across various sociopolitically charged topics. This quantitative assessment revealed discernible disparities in the models' ability to differentiate between neutral summaries and those with explicit political biases. 

For example, in the context of Abortion, the accuracies for distinguishing between default and Democrat-aligned summaries ranged from 52.3\% to 66.83\%, whereas for Republican-aligned summaries, this range was from 53.26\% to 69.21\%. These values indicate a more substantial differentiation in summaries leaning towards Republican ideologies, a pattern that was consistent across other topics as well. Notably, \texttt{PaLM-2} frequently demonstrated the highest \(\textit{diff}(.)\) values, signifying its greater deviation from neutral baselines and suggesting a more pronounced ideological bias within its summaries. Across models and topics, we observe that political biases manifest within LLM-generated content, revealing both the capabilities and the limitations of these models in maintaining neutrality in ideologically sensitive discourse.

\subsubsection{Algorithmic Monoculture Quantification using Classifiers: Do they use the same language?}

\begin{figure}[h!]
    \centering
    \begin{subfigure}[b]{0.49\textwidth}
        \centering
        \includegraphics[width=\textwidth]{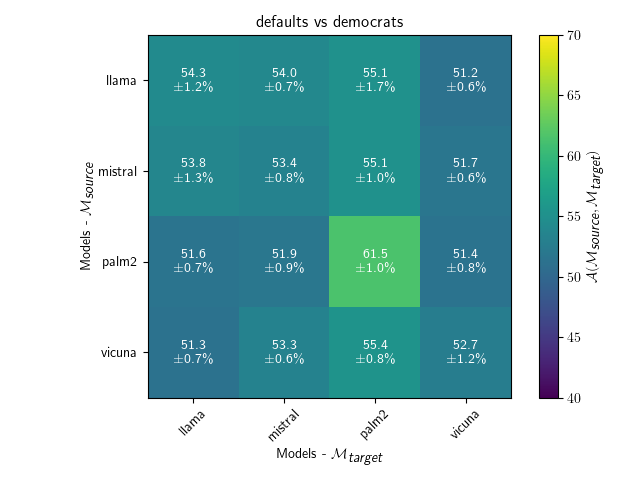}
        \caption{\scriptsize Heatmap of classifier performance (\(\mathcal{A}\) scores) for distinguishing between default and Democrat-aligned summaries across different LLMs (\(\mathcal{M}_{\textit{source}}\) vs. \(\mathcal{M}_{\textit{target}}\)).}
        \label{fig:democratclfmonoculture}
    \end{subfigure}
    \hfill
    \begin{subfigure}[b]{0.49\textwidth}
        \centering
        \includegraphics[width=\textwidth]{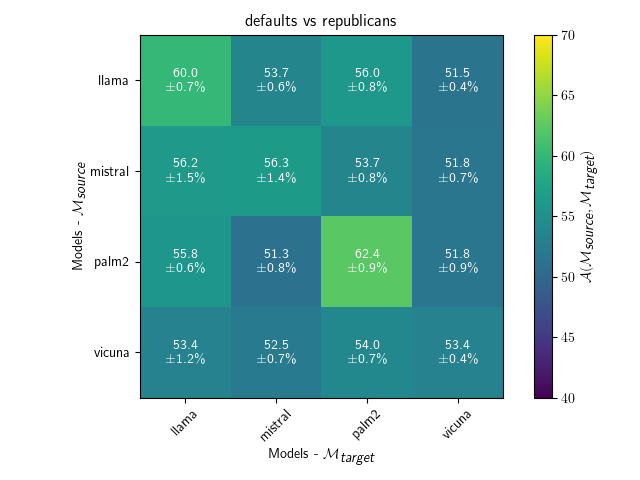}
        \caption{\scriptsize Heatmap of classifier performance (\(\mathcal{A}\) scores) for distinguishing between default and Republican-aligned summaries across different LLMs (\(\mathcal{M}_{\textit{source}}\) vs. \(\mathcal{M}_{\textit{target}}\)).}
        \label{fig:republicanclfmonoculture}
    \end{subfigure}
    \caption{Comparative heatmaps illustrating classifier performance (\(\mathcal{A}\) scores) in identifying political biases within LLM-generated summaries. These visualizations reveal the degree of transferability of ideological biases and linguistic strategies, providing insights into the prevalence of algorithmic monoculture.}
    \label{fig:clfmonoculture}
\end{figure}

We analyzed the transferability of ideological biases across LLMs by applying SentenceTransformer-based classifiers trained on one model's summaries (\(\mathcal{M}_{\textit{source}}\)) to those generated by another (\(\mathcal{M}_{\textit{target}}\)). This cross-LLM evaluation, denoted as \(\mathcal{A}(\mathcal{M}_{\textit{source}}, \mathcal{M}_{\textit{target}})\), aimed to unravel the extent of shared ideological biases and linguistic strategies among different models. Our findings, depicted through heatmaps in Figure~\ref{fig:clfmonoculture}, reveal nuanced insights into the ideological alignment and diversity within the LLM ecosystem.

On average, the \(\mathcal{A}\) score for classifiers differentiating between default and Democrat-aligned summaries, excluding self-referential evaluations (\(\mathcal{M}_{\textit{source}} \neq \mathcal{M}_{\textit{target}}\)), stood at 52.98\%. Similarly, for default versus Republican-aligned summaries, the average \(\mathcal{A}\) score was slightly higher at 53.47\%. In contrast, when classifiers were applied to the same model they were trained on (\(\mathcal{M}_{\textit{source}} = \mathcal{M}_{\textit{target}}\)), the scores increased to 55.45\% for Democrat-default differentiation and 57.99\% for Republican-default differentiation. Even so, the presence of this monoculture is further nuanced by the observation that despite a decrease in classification accuracy when classifiers are applied across different LLMs, the scores remain significantly close to \(\mathcal{M}_{\textit{source}} = \mathcal{M}_{\textit{target}}\) scores, illustrating that certain ideological biases and linguistic strategies may indeed be transferable across models. This subtlety suggests that while no model is entirely neutral or unbiased, there exists a degree of commonality in how these biases are manifested, hinting at underlying patterns of monoculture that span across different LLMs.

\section{Discussion and Analysis}

\begin{table*}[h!]
    \centering
    \scriptsize
    \begin{tabular}{|l|r|r|r|r|r|}
    \hline
    \textbf{Topics / Models} & \textbf{LLaMA-7B (\%)} & \textbf{Mistral-7B (\%)} & \textbf{PaLM-2 (\%)} & \textbf{Vicuna-7B (\%)} & \textbf{Averages (\%)} \\
    \hline
    \textbf{Abortion}      & -4.30\%    & \textbf{-3.51\%}      & -2.39\%   & -0.96\%     & -2.79\%    \\
    \textbf{Gun Control/Rights}   & -5.21\%    & -1.14\%      & \textbf{-9.49}\%   & \textbf{1.33\%}     & \textbf{-3.63\%}    \\
    \textbf{Healthcare}    & \textbf{-6.14\%}    & -2.83\%      & -4.14\%   & 0.75\%     & -3.09\%    \\
    \textbf{Immigration}   & -5.66\%    & -2.95\%      & -0.89\%   & -0.79\%     & -2.57\%    \\
    \textbf{LGBTQ+}         & -2.98\%    & -1.87\%      & -2.83\%   & -0.53\%     & -2.05\%    \\ \hline
    \textbf{Averages}      & -4.86\%    & -2.46\%      & -3.95\%   & -0.04\%     & -2.83\%    \\
    \hline
    \end{tabular}
    \caption{This table presents the Polarization Index ($\mathcal{P}$) for five sociopolitical topics, comparing across language models (LLMs). Negative values signal a bias towards Democrat perspectives whereas positive values signal a bias towards Republican perspectives, illustrating how both topics and LLMs vary in ideological alignment. The bolded elements highlight the minimum/maximum values in each column, emphasizing the greatest polarization bias across both topics and language models.}
    \label{tab:polarizationfull}
\end{table*}

In our analysis of ideological biases across LLMs, we uncover intriguing patterns of polarization on five sociopolitical topics, as illustrated in our detailed comparison table, Table~\ref{tab:polarizationfull}. Notably, \texttt{PaLM-2} exhibits a striking -9.49\% Polarization Index on Gun Control/Rights, the most pronounced bias across all models and topics, highlighting its distinct skew towards Democrat perspectives. Conversely, \texttt{Vicuna-7B} shows an unexpected positive tilt with a 1.33\% on the same topic, suggesting a nuanced Republican alignment. Overall among models, \texttt{LLaMA-7B} carries the highest average polarization at -4.86\%, indicating a stronger, consistent pro-democratic bias.

\begin{table}[h!]
\centering
\scriptsize
\begin{tabular}{|lcc|}
\toprule
\textbf{Topic} & \textbf{Mean Polarization (\%)} & \textbf{Max Polarization (\%)} \\
\midrule
Abortion & -2.79 & -4.30 \\
Gun Control/Rights & -3.63 & -9.49 \\
Healthcare & -3.09 & -6.14 \\
Immigration & -2.57 & -5.66 \\
LGBTQ+ & -2.05 & -2.98 \\
\bottomrule
\end{tabular}
\caption{This table showcases the mean and maximum Polarization Index ($\mathcal{P}$) values across the five sociopolitical topics explored in the paper. A negative score throughout the table indicates a uniform bias towards Democrat-aligned perspectives. The variation in polarization highlights the differential ideological tilt presented in LLM-generated summaries.}
\label{tab:polarizationcomparison}
\vspace{-0.1cm}
\end{table}

The comprehensive analysis of ideological biases inherent in LLMs elucidates significant disparities in vocabulary usage and the ability of models to generate politically aligned summaries. We observe that a significant portion of the vocabulary presented in Table~\ref{tab:divergingvocab} is closely associated with the key talking points of each political party. The Vocabulary Analysis and Polarization Quantification sections collectively paint a picture of how deeply and in what ways these biases are embedded within LLM outputs. Particularly, the findings surrounding the Polarization Index ($\mathcal{P}$) across various sociopolitically charged topics, as outlined in the Comparative Analysis of Mean and Maximum Polarization. Table~\ref{tab:polarizationcomparison} offers a window into the nuanced ideological landscapes these models navigate.

The consistently negative Polarization Indices across all topics suggest a systematic bias toward Democrat-aligned viewpoints in LLM-generated summaries. This indicates not just an inherent leaning within the models but also highlights the subtleties of ideological representation within machine-generated texts. Gun Control/Rights, exhibiting the highest mean and maximum polarization, underscores the significant ideological divergence in how this topic is framed by the models. In contrast, the relatively lower polarization in LGBTQ+ topics suggests a more nuanced or balanced representation, albeit still tilted towards Democrat viewpoints. The implications of such biases become particularly significant in the context of upcoming U.S. elections.

Moreover, the exploration of algorithmic monoculture revealed through \(\mathcal{A}(\mathcal{M}_{\textit{source}}, \mathcal{M}_{\textit{target}})\) scores and the Vocabulary Analysis further complicates the picture. There seem to be a detectable level of commonality in biases across these different models, as explored through vocabulary overlap and classification tasks, which suggests that while LLMs may differ in architecture and organizational control, they may be converging towards certain ideological biases.

\section{Conclusion}

Our analysis was not merely an exploration into whether LLMs can be made to regurgitate political ideologies, but rather an investigation into whether seemingly neutral tasks, such as text summarization, can inadvertently skew towards a particular political ideology. The Polarization Index (P) values presented in Table \ref{tab:polarizationcomparison} reveal a consistent bias towards Democrat-aligned perspectives across all five sociopolitical topics examined, with a maximum polarization score of -9.49\%. This suggests that even when tasked with objective summarization, LLMs may introduce a subtle political slant in their outputs. Further complicating this issue is our investigation into the concept of algorithmic monoculture, which demonstrates that this is not an isolated phenomenon; a significant overlap in ideological vocabulary across models indicates a broader pattern of bias. Specifically, the overall average $\mathcal{CI}$ for Democrat-leaning vocabulary stands at 55.37\%, pointing towards a consensus among models on Democrat-biased terms. These findings suggest that the observed biases are not unique to a single LLM but are rather indicative of a wider algorithmic tendency to favor certain ideological perspectives.

The implications of these findings are particularly significant in the context of the upcoming U.S. elections. Traditional media outlets, such as newspapers, television channels, and search engines, are expected to maintain political neutrality to ensure a fair and unbiased information landscape. However, as LLMs increasingly become a go-to resource for tasks like news synopsis, key point extraction, and even as potential replacements for search engines, their inherent biases could lead to the manipulation of public opinion and, consequently, election outcomes. Moreover, the growing use of LLMs as direct knowledge sources by younger generations raises concerns about the potential for warped political perceptions. If these models consistently expose users to one political sphere more frequently than others, it could lead to the formation of echo chambers and the reinforcement of biased political views. Given the critical role that information plays in shaping public discourse and electoral processes, it is imperative that the development and deployment of LLMs prioritize political neutrality. In conclusion, our analysis underscores the need for a proactive approach to managing the ideological biases present in LLM-generated content.

\section{Ethics Statement}

In undertaking this analysis of ideological biases within Large Language Models (LLMs), we were guided by a commitment to ethical rigor. The investigation into how these advanced technologies may influence societal narratives and potentially contribute to polarization necessitates careful ethical consideration. Our research adheres to principles of transparency and accountability, ensuring that our methodologies are both reproducible and grounded in an understanding of their broader societal implications.

\subsection{Limitations}

A notable limitation of our study is the exclusion of some of the larger models like GPT-4 or GPT-3.5, primarily due to the expansive scale of our experiment which involved generating and analyzing over 240k summaries. While we did include \texttt{PaLM-2} —a model of considerable size that is freely available and thus accessible for our analysis—the decision to focus on smaller 7B models such as \texttt{LLaMA-7b, Mistral-7B}, and \texttt{Vicuna-7B} was driven by considerations of computational feasibility. It is important to note that these models, despite their smaller size, offer substantial insights into the potential biases present in LLM outputs. However, they may not fully capture the breadth of ideological biases that larger models, with their more extensive training datasets and potentially more nuanced understanding, could exhibit.

\subsection{Ethical Considerations \& Future Work}

Acknowledging this, we see the analysis of larger models as an essential avenue for future research. Expanding our investigation to include these models will allow for a more comprehensive assessment of ideological biases across a broader spectrum of AI technologies. This future work will further our understanding of how model scale influences the manifestation of biases, contributing valuable insights to discussions on the ethical deployment of LLMs in various societal contexts.

Our research underscores the need for ongoing dialogue within the AI community and among stakeholders about the ethical use of LLMs, especially as they become more embedded in societal infrastructures and processes. By carefully considering the implications of our findings and acknowledging the limitations of our current study, we contribute to the responsible advancement of AI technologies that respect and uphold democratic values and diversity.

\bibliographystyle{plainnat} 
\bibliography{noiseaudit.bib}

\end{document}